\pdfoutput=1

\documentclass[11pt]{article}

\usepackage{emnlp2021}

\usepackage{times}
\usepackage{latexsym}
\usepackage{paralist}

\usepackage[T1]{fontenc}
\usepackage[utf8]{inputenc}

\usepackage{microtype}

\usepackage{graphicx}
\usepackage{amsmath}
\usepackage{amsfonts}
\usepackage[ruled]{algorithm2e}
\usepackage{booktabs}
\usepackage{multirow}
\usepackage{subfigure}
\usepackage{xcolor}
\usepackage{balance}

\title{Knowing False Negatives: An Adversarial Training Method for\\ Distantly Supervised Relation Extraction}

\author{Kailong Hao$^\dagger$ \quad 
        Botao Yu$^\dagger$ \quad 
        Wei Hu$^{\dagger,\,\ddagger,\,}$\thanks{\,\, Corresponding author} \\
	$^\dagger$ State Key Laboratory for Novel Software Technology, Nanjing University, China \\
	$^\ddagger$ National Institute of Healthcare Data Science, Nanjing University, China \\
	\texttt{klhao.nju@gmail.com, btyu.nju@gmail.com, whu@nju.edu.cn} 
}

\begin{document}
\maketitle
\begin{abstract}
Distantly supervised relation extraction (RE) automatically aligns unstructured text with relation instances in a knowledge base (KB). 
Due to the incompleteness of current KBs, sentences implying certain relations may be annotated as N/A instances, which causes the so-called false negative (FN) problem. 
Current RE methods usually overlook this problem, inducing improper biases in both training and testing procedures. 
To address this issue, we propose a two-stage approach. First, it finds out possible FN samples by heuristically leveraging the memory mechanism of deep neural networks. 
Then, it aligns those unlabeled data with the training data into a unified feature space by adversarial training to assign pseudo labels and further utilize the information contained in them. 
Experiments on two wildly-used benchmark datasets demonstrate the effectiveness of our approach.
\end{abstract}

\section{Introduction}

Relation extraction (RE), defined as the task of identifying relations from unstructured text given two entity mentions, is a key component in many NLP applications such as knowledge base (KB) population \cite{knowledge-base-population,DeepDive-knowledge-base-construction} and question answering \cite{yu-etal-2017-improved}. 
Supervised RE requires a large amount of human labeled data, which is often labor-intensive and time-consuming. Distant supervision (DS), proposed by \cite{mintz}, deals with this problem by aligning a large corpus with a KB such as Freebase to provide weak supervision. It relies on the assumption that \emph{if two entities participate in a relation in a KB, then any sentence containing this entity pair expresses this relation}.

However, this assumption is too strong and brings in noises, particularly when the training KB is an external information source and not primarily derived from the training corpus. To alleviate this problem, \citet{riedel} relax this assumption to \emph{if two entities participate in a relation, \textbf{at-least-one} sentence that mentions these two entities expresses this relation}.
\citet{MIL4RE2007} propose multi-instance learning by organizing sentences with the same entity pair into one set, referred to as a \textit{bag}. Following this setting, \citet{riedel, Hoffmann, Surdeanu} propose diverse hand-crafted features. \citet{zeng-etal-2014-relation, PCNN} leverage convolutional neural networks (CNN) to learn the representations of instances. \citet{ATT} apply an attention mechanism to select informative sentences from a bag. Recently, graph convolutional networks (GCN) have been effectively employed for capturing the syntactic information from dependency trees \cite{RESIDE}.

\begin{table}
\centering
\resizebox{\columnwidth}{!}{
\begin{tabular}{p{6.5cm}c}
\toprule
\textbf{Sentences} & \textbf{Relations}\\
\midrule
\textcolor{orange}{\textbf{[Romano Mussolini]}} , the Italian jazz pianist , ... , died on Tuesday in \textcolor{orange}{\textbf{[Rome]}} . & death\_place \\
\midrule
... in the \textcolor{orange}{\textbf{[Uptown]}} neighborhood of \textcolor{orange}{\textbf{[New Orleans]}} , ... , where it was founded . & neighborhood \\
\bottomrule
\end{tabular}}
\caption{False negative examples excerpted from the NYT10 \cite{riedel} N/A set.}
\label{tab:NA}
\end{table}

Most of the above works focus on the false positive (FP) problem (which is caused by the strong DS assumption), but totally neglect the false negative (FN) problem, which is also important and induces improper biases in both training and testing procedures. 
DS treats a sentence as a negative sample (marked as N/A) whose entity pair does not have a known relation in the KB. However, due to the incompleteness of current KBs, sentences implying predefined relations are likely to be mislabeled as N/A. For example, over 70\% of people in Freebase do not have birth places \cite{Dong-Xin-incompleteness}. 
As shown in Table \ref{tab:NA}, we have annotated the ground-truth relations for two FN sentences. In fact, the missing facts in KBs yield plenty of FN sentences in the automatically annotated datasets.

Generative adversarial networks (GAN) \cite{goodfellow2014generative} is first introduced into RE by \cite{qin-etal-2018-dsgan} to learn a sentence-level FP filter. However, they only focus on the FP problem and their model cannot be generalized to other scenarios. 
\citet{GANDriven} attempt to solve the FN problem using entity descriptions from Wikipedia to filter FN samples from the N/A set and further utilize GAN in a semi-supervised manner. 
However, their method heavily depends on external resources and is not applicable to all situations.
Moreover, their filtering heuristic ``\emph{two entities mention each other on their Wikipedia pages may imply the predefined relations}'' is even stronger than that of \cite{riedel}. It may filter some noisy data but at the same time introducing even more noises. For example, in the sentence ``\textit{... to the poetry from [India], China and [Korea] that stretches the book ...}'' excerpted from NYT10, India and Korea mention each other on Wikipedia, but this sentence implies no relation.

In this paper, we propose a novel two-stage approach for distantly supervised RE. 
In the first stage, called \textbf{mining}, we find out possible FN samples from the N/A set by heuristically leveraging the memory mechanism of deep neural networks. According to \cite{memory}, while deep neural networks are capable of memorizing noisy data, they tend to preferentially learn simple patterns first, that is to say, deep neural networks tend to learn and memorize patterns from clean instances within a noisy dataset. We design a Transformer-based \cite{AttentionisAll} deep filter to mine FN samples from the N/A set. 
In the second stage, called \textbf{aligning}, we formulate the problem into a domain adaptation (DA) paradigm. We exploit a gradient reversal layer (GRL) to align the mined unlabeled data from stage one with the training data into a unified feature space. After aligning, each sentence is assigned with a pseudo label and a confidence score, which provide extra information and attenuate incorrect biases in both training and testing procedures. 

In summary, our main contributions are fourfold:

\begin{compactitem}
    \item We propose a simple yet effective method to filter noises in a DS dataset by leveraging the memory mechanism of deep neural networks, without any external resources.
    
    \item We formulate distantly supervised RE as a DA paradigm and utilize adversarial training to align unlabeled data with training data into a unified space, and generate pseudo labels to provide additional supervision.
    
    \item We achieve new state-of-the-art on two popular benchmark datasets NYT10 \cite{riedel} and GIDS \cite{GIDS}.
    
    \item By mining the test set, we show that the FN problem greatly misleads the evaluation of DS models and deserves further study.
\end{compactitem}

\section{Related Work}

\paragraph{Distant supervision.} Supervised RE requires a large amount of human-labeled training data, which is labor-intensive and time-consuming. To address this limitation, \citet{mintz} propose DS by heuristically aligning a text corpus to a KB. \citet{riedel} relax DS for multi-instance single-label learning. Subsequently, to handle the overlapping relations between entity pairs, \citet{Hoffmann,Surdeanu} propose the multi-instance multi-label learning paradigm. Currently, DS is already a common practice in RE.

\paragraph{Neural relation extraction.} The above works strongly rely on the quality of hand-engineered features. \citet{zeng-etal-2014-relation} first propose an end-to-end CNN-based neural network to automatically capture relevant lexical-level and sentence-level features. \citet{PCNN,ATT} further improve this through piecewise max pooling and selective attention \cite{bahdanau2016neural}. 
\citet{zhou-etal-2016-attention-based} propose an attention-based LSTM to capture the most important semantic information in sentences. External knowledge like entity descriptions and type information has been used for RE \cite{yaghoobzadeh-etal-2017-noise,RESIDE}. Pre-trained language models contain a notable amount of semantic information and commonsense knowledge, and several works have applied them to RE \cite{DISTRE,Xiao_Tan_Fan_Xu_Zhu_2020}.

\paragraph{Adversarial training.} Adversarial training is a machine learning technique that improves the networks using an adversarial objective function or deceptive samples. \citet{wu-etal-2017-adversarial} bring it in RE by adding adversarial noises to the training data. \citet{qin-etal-2018-dsgan} propose DSGAN to learn a generator that explicitly filters FP instances from the training dataset. \citet{GANDriven} propose a semi-distant supervision method by first splitting a dataset through entity descriptions and then using GAN to make full use of unsupervised data. \citet{RDSGAN} learn the distribution of true positive instances through adversarial training and select valid instances via a rank-based model.

\paragraph{Learning with noisy labels.} As noisy labels degrade the generalization performance of deep neural networks, learning from noisy labels (a.k.a. robust training) has become an important task in modern deep learning \cite{song2021noisylabel}. \citet{memory} find out that, although deep networks are capable of memorizing noisy data, they tend to learn simple patterns first. Co-teaching \cite{Coteaching} trains two deep neural networks simultaneously and lets them teach each other given every mini-batch. \citet{Robustsemi, SELF, li2020dividemix} transform the problem into a semi-supervised learning task by treating possibly false-labeled samples as unlabeled.

\section{Methodology}
In this section, we introduce our two-stage framework called FAN (False negative Adversarial Networks) in detail. First, we describe how we discover possibly wrong-labeled sentences from the N/A set. Then, we introduce our adversarial DA method, which assigns pseudo labels to unlabeled data with confidence scores.

\subsection{Stage \uppercase\expandafter{\romannumeral1}: Mining}
We define a distantly supervised dataset $\mathcal{D} = \{s_1,s_2,\dots,s_N\}$, where each sample $s_i$ is a quadruple, consisting of an input sequence of tokens $t_i = [t_i^1, \dots, t_i^n]$, $head_i$ and $tail_i$ for head and tail entity positions in sequence $t_i$, respectively, and the corresponding relation $r_i$ assigned by DS. 
We split $\mathcal{D}$ into two parts: sentences with predefined relations are divided into the positive set, denoted by $\mathcal{P}$; sentences implying no relations are divided into the negative set, denoted by $\mathcal{N}$, $\mathcal{D} = \mathcal{P} \cup \mathcal{N}$. In this work, we focus on the noises in $\mathcal{N}$, where sentences may be mislabeled by incomplete KBs and useful information is not yet fully discovered.

According to \cite{memory}, deep neural networks are prone to learn clean samples first, and then gradually learn noisy samples. Following \cite{Decoupling,MentorNet,Coteaching,SELF}, after proper training, we filter samples in $\mathcal{N}$ with logits larger than threshold $\theta$ as possible FN samples. All FN samples form a set $\mathcal{M}$ and remaining samples form a set $\mathcal{N^\prime}$, where $\mathcal{N} = \mathcal{N^\prime} \cup \mathcal{M}$. The original training dataset $\mathcal{D}$ is refined to $\mathcal{D^\prime}$, where $\mathcal{D^\prime} = \mathcal{P} \cup \mathcal{N^\prime}$. The deep noise filter can capture meaningful semantic patterns to differentiate between $\mathcal{P}$ and $\mathcal{N}$. For the samples in $\mathcal{N}$ with logits larger than $\theta$, they may imply predefined relations but only the DS annotations are inaccurate. We argue that $\mathcal{M}$ can be considered as unlabeled data in a semi-supervised way to provide supplementary information.

\subsubsection{Sentence Encoder}
\label{sec:sentence-encoder}

\begin{figure}
    \centering
    \includegraphics[width=\columnwidth]{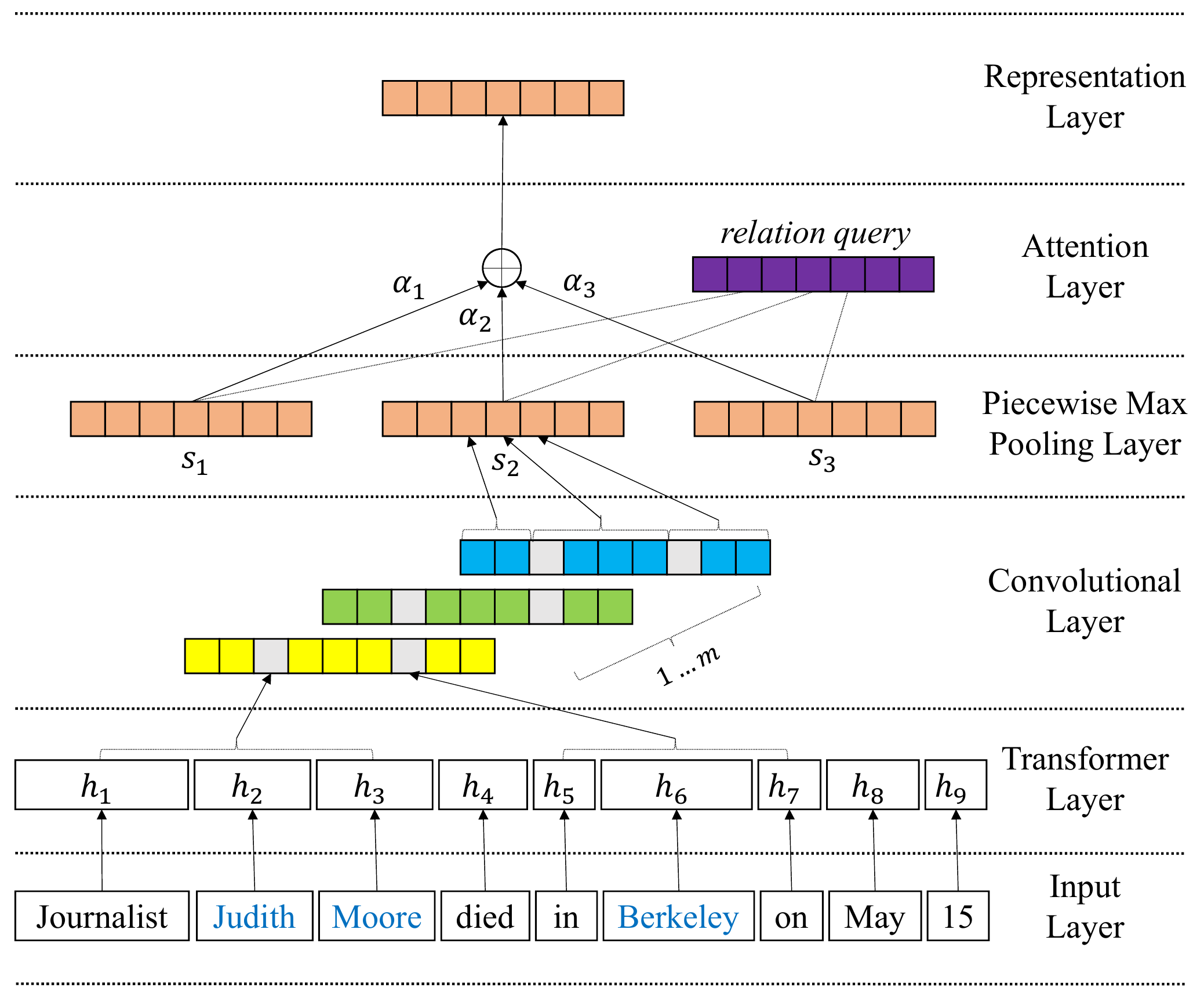}
    \caption{Architecture of encoder.}
    \label{fig:encoder}
\end{figure}

As for the mining step, we take a mapping function $f(\cdot)$ to map $r_i$ to a binary label. If $r_i$ is N/A, $f(r_i) = 0$; otherwise $f(r_i) = 1$. We use pre-trained language model BERT \cite{BERT} as our embedding module, as it contains a large amount of semantic information and commonsense knowledge. Token sequence $t_i = [t_i^1, \dots, t_i^n]$ is fed into the pre-trained model and the last hidden representation $\mathbf{h}_i = [\mathbf{h}_i^1, \dots, \mathbf{h}_i^n]$ is used as our token embedding.

CNN is a widely used architecture for capturing local contextual information \cite{zeng-etal-2014-relation}. The convolution operation involves taking the dot product of the convolutional filter $\mathbf{W}$ with each \textit{$k$-gram} in the sequence $\mathbf{h}_i$ to obtain a new representation $\mathbf{p}_i$:
\begin{equation}
    p_i^j = \mathbf{W} \cdot [\mathbf{h}_{i}^{j-k+1}:\ldots:\mathbf{h}_i^{j}],
\end{equation}
where $p_i^j$ is the $j$-th dimension of $\mathbf{p}_i$. $\mathbf{W} \in \mathbb{R}^{k \times d}$ is the convolutional filter, where $k$ is the kernel size and $d$ is the hidden dimension. $[\mathbf{h}_{i}^{j-k+1}:\ldots:\mathbf{h}_i^{j}]$ refers to the concatenation from $\mathbf{h}_{i}^{j-k+1}$ to $\mathbf{h}_i^{j}$. In order to capture diverse features and extract the local information at different levels, we make use of multiple filters and varied kernel sizes. 

In RE, representation $\mathbf{p}_i$ can be partitioned into three parts according to $head_i$ and $tail_i$, i.e., $\mathbf{p}_i = \{\mathbf{p}_{i_1}, \mathbf{p}_{i_2}, \mathbf{p}_{i_3}\}$. To capture the structural information between two entities and obtain fine-grained features, we take piecewise max pooling as \cite{PCNN}. For each convolutional filter, we can obtain a three-dimensional vector $\mathbf{q}_i$:
\begin{equation}
    \mathbf{q}_i = [\max (\mathbf{p}_{i_1}), \max (\mathbf{p}_{i_2}), \max (\mathbf{p}_{i_3})].
\end{equation}

We concatenate vectors from multiple convolutional filters and get the sentence representation $\mathbf{v}_i \in \mathbb{R}^{3m}$, where $m$ is the number of convolutional filters with varied kernel sizes. The architecture of the sentence encoder is shown in Figure \ref{fig:encoder}.

\subsubsection{Noise Filter}
Through sentence encoding, each sample $s_i = (t_i, head_i, tail_i, r_i)$ is transformed into a fixed-dimensional vector $\mathbf{v}_i \in \mathbb{R}^{3m}$. 
We feed it into a multilayer perceptron to obtain the probability:
\begin{equation}
    o_i = \sigma(\mathbf{W}_f\mathbf{v}_i + b_f),
\end{equation}
where $\mathbf{W}_f \in \mathbb{R}^{1 \times 3m}$ is the transformation matrix and $b_f$ is the bias. $\sigma(\cdot)$ denotes the sigmoid activation. Output $o_i$ implies the probability of current sentence belonging to $\mathcal{P}$. 

We use the binary cross-entropy as our loss to train the deep noise filter in an end-to-end way:
\begin{align}
\resizebox{.86\columnwidth}{!}{$
    \begin{aligned}
        L_{f} =\mathbb{E}_{s_i \sim \mathcal{D}} \big[&- f(r_i)\log(\mathrm{P}(s_i)) \\ 
           &- (1-f(r_i))\log(1-\mathrm{P}(s_i))\big].
    \end{aligned}
$}
\end{align}

\subsection{Stage \uppercase\expandafter{\romannumeral2}: Aligning}
\begin{figure*}[htbp]
    \centering
    \includegraphics[width=\textwidth]{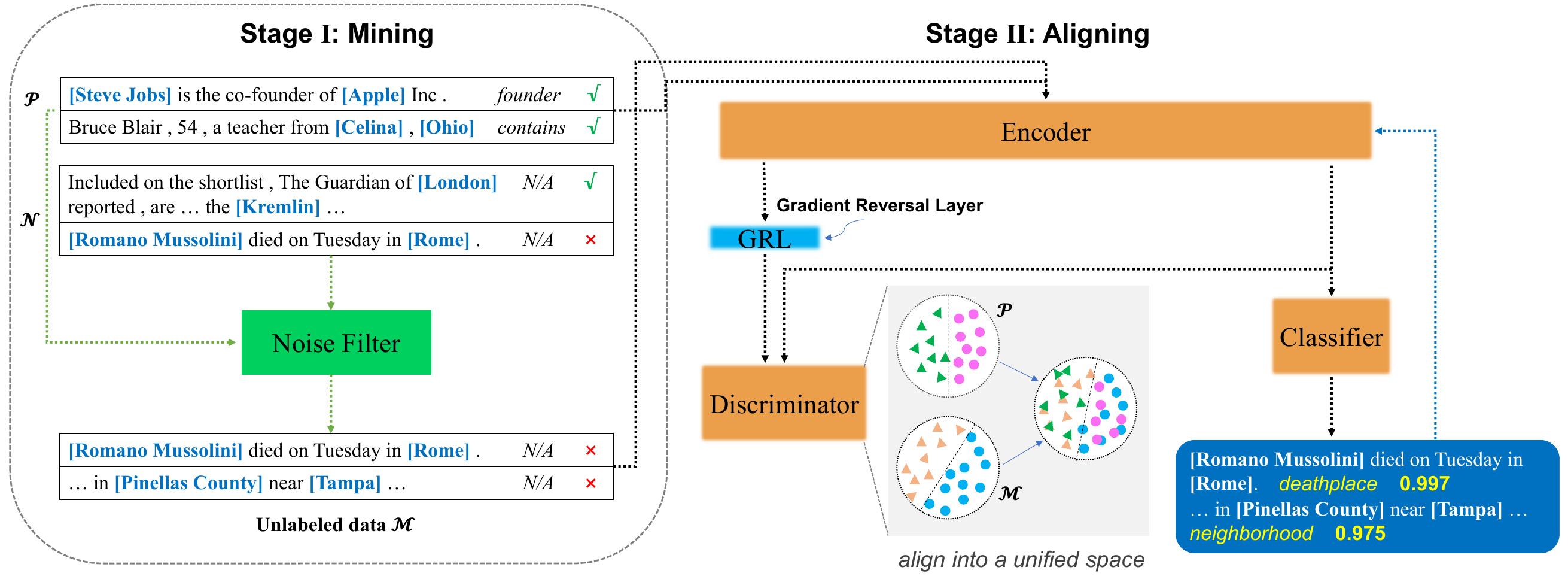}
    \caption{Overview of FAN. Informative unlabeled samples are first mined through the noise filter and then fed into the adversarial module to generate pseudo labels with confidences. The adjusted training data are used for final training.}
    \label{fig:Overview}
\end{figure*}

Instances in $\mathcal{M}$ are mislabeled due to the incompleteness of KBs and the distribution may be different from the original training dataset \cite{ye2019looking}.
A naive way is just dropping those data and using adjusted dataset $\mathcal{D^\prime}$ for training. Doing like this would lose useful information contained in $\mathcal{M}$ and thus is not optimal. Unlabeled data can be annotated by humans, but it is time-consuming and is not applicable to large datasets. 

In fact, these unlabeled samples imply predefined relations and can be used together with $\mathcal{D^\prime}$ in a semi-supervised learning paradigm. We formulate this problem as a DA task and the objective is aligning the distributions of $\mathcal{M}$ and $\mathcal{P}$ into a unified feature space. 
To achieve this objective, we propose a method inspired by GAN. The generator tries to fool the discriminator so that it cannot distinguish the samples in $\mathcal{M}$ and $\mathcal{P}$. On the contrary, the discriminator tries its best to differentiate them. The training procedure forms a classic min-max game by adversarial objective functions. The overall architecture is shown in Figure \ref{fig:Overview}.

\subsubsection{Bag Encoder}
The sentence encoding layer reuses the architecture in Section  \ref{sec:sentence-encoder}. 
Due to the noisy DS annotations, multi-instance learning is introduced and relation classification is applied on the bag level. 
One bag $\mathcal{B} = \{s_1,\dots,s_t\}$ contains $t$ sentences for the same entity pair. Bag representation $\mathbf{g}_i$ is derived from a weighted sum over the individual sentence representations:
\begin{equation}
    \mathbf{g}_i = \sum_j \alpha_j\mathbf{v}_j,
\end{equation}
where $\alpha_j$ is the weight assigned to the corresponding sentence computed through selective attention \cite{ATT}. The weight is obtained by computing the similarity between the learned relation query representation $\mathbf{r}_i \in \mathbb{R}^{3m}$ and each sentence:
\begin{equation}
    \alpha_i = \frac{\exp(\mathbf{v}_i \cdot \mathbf{r}_i)}{\sum_{j=1}^t \exp(\mathbf{v}_j \cdot \mathbf{r}_i)}.
\end{equation}

\subsubsection{Relation Classifier}
For a bag in $\mathcal{P}$, its DS label is known. To compute the probability distribution over relations, a linear layer followed by a softmax layer is applied to a bag representation $\mathbf{g}_i \in \mathbb{R}^{3m}$:
\begin{equation}
    \mathrm{P}(r_i \,|\, \mathbf{g}_i) = \mathrm{softmax} (\mathbf{W}_c \mathbf{g}_i + \mathbf{b}_c),
\end{equation}
where $\mathbf{W}_c \in \mathbb{R}^{l \times 3m}$, and $l$ is equal to the number of predefined relations. 

During training, we aim to optimize the following cross-entropy loss:
\begin{equation}
    L_{cls} = \mathbb{E}_{\mathbf{g}_i \sim \mathcal{P}} \big[-\log \mathrm{P}(r_i\,|\,\mathbf{g}_i)\big].
    \label{eq:lcls}
\end{equation}


\subsubsection{Gradient Reversal Layer}
Pre-trained language models have shown great power in many NLP tasks. They are huge in size and have tremendous ability to fit distributions. Following \cite{DANN, ADAN}, we use a GRL after the bag encoder. When forward passing, it works as an identity function, while when back propagating, it reverses gradients to their opposite:
\begin{equation}
    \mathrm{GRL} = \mathbb{I}(\cdot)\nabla_\Theta,
\end{equation}
where $\Theta$ denotes parameters of the bag encoder. When forward passing, $\mathbb{I}(\cdot) = 1$, and when back propagating, $\mathbb{I}(\cdot) = -1$.

\subsubsection{Discriminator}
Given a bag representation $\mathbf{g}_i \in \mathbb{R}^{3m}$, the discriminator first implements an affine transformation and then uses sigmoid function $\sigma(\cdot)$ to obtain the probability distribution: 
\begin{equation}
    o_i^\prime = \sigma(\mathbf{W}_d\mathbf{g}_i + b_d).
\end{equation}

\subsection{Training Objective}
We use adversarial training to generate a unified data distribution. To enforce instances of the same class closer and push away instances from different classes, a contrastive loss is designed for better feature representation.

\subsubsection{Adversarial Loss}
The bag encoder is optimized to give separate representations for instances in $\mathcal{P}$, so that samples from different classes can be easily distinguished by the relation classifier. In the meantime, it forces the distribution of $\mathcal{M}$ to fit into the distribution of $\mathcal{P}$. 
The encoder here plays two roles: representation learner and distribution adapter. The classification learning objective is Eq.~(\ref{eq:lcls}), and the generator objective is
\begin{equation}
    L_g = -\mathbb{E}_{s \sim \mathcal{M}} \big[\mathrm{D}(\mathrm{G(s)})\big].
\end{equation}

On the contrary, the discriminator attempts to distinguish samples from $\mathcal{M}$ with $\mathcal{P}$. 
Straightforwardly, for generator, the labels of samples in $\mathcal{M}$ are $1$; but for discriminator, the labels of samples in $\mathcal{M}$ are $0$. 
The labels of instances in $\mathcal{P}$ are always $1$. The discriminator objective is
\begin{equation}
\resizebox{.86\columnwidth}{!}{$
    L_d = -\mathbb{E}_{s_i \sim \mathcal{P}} \big[\mathrm{D}(s_i)\big] 
    + \mathbb{E} _{s_j \sim \mathcal{M}}\big[\mathrm{D}(\mathrm{G}(s_j))\big].
$}
\end{equation}

Generator and discriminator improve each other in iterations. 

\subsubsection{Contrastive Loss}
Bag representations are expected to be able to cluster instances with the same relation. We aim to increase the distances between samples with different relations and reduce the variance of distances with the same relation.

Simply put, given two instances, their similarity score should be high if they belong to the same relation and low otherwise. We use the contrastive loss \cite{ContrastiveLoss} for this objective. Following \cite{NERO}, given a bag representation $\mathbf{g}$, we divide all the other instances with the same relation type as $\mathcal{Q}_+$ and the ones with different types as $\mathcal{Q}_-$. 
The contrastive loss can be formulated as 
\begin{align}
    L_{ctra} = \mathbb{E}_{\mathbf{g} \sim \mathcal{P}} \Big[& \max \limits_{\mathbf{g}_i \in \mathcal{Q}_{+}} \mathrm{dist}_{+} (\mathbf{g}, \mathbf{g}_i) \notag \\
    & - \min \limits_{\mathbf{g}_j \in \mathcal{Q}_{-}} \mathrm{dist}_{-}(\mathbf{g}, \mathbf{g}_j)\Big],
\end{align}
where the measurement of distance is defined as follows:
\begin{align}
\resizebox{.86\columnwidth}{!}{$
	\begin{aligned}
     \mathrm{dist}_{+}(\mathbf{g}, \mathbf{g}_i) &= \big(\max (\tau - \cos (\mathbf{g}, \mathbf{g}_i), 0)\big)^2, \\
     \mathrm{dist}_{-}(\mathbf{g}, \mathbf{g}_j) &= 1 - \big(\max (\cos (\mathbf{g}, \mathbf{g}_j), 0)\big)^2,
    \end{aligned}
$}
\end{align}
where $\tau$ is a hyperparameter for avoiding collapse of the representations of bags, $\cos(\cdot)$ denotes the cosine function.

\subsubsection{Overall Loss}
The adversarial training procedure is modeled as multi-task learning and trained in an end-to-end way.
The overall objective function is
\begin{equation}
    L = L_{cls} + \alpha \cdot L_{g} + \beta \cdot L_{d} + \gamma \cdot L_{ctra},
\end{equation}
where $\alpha, \beta, \gamma$ are hyperparameters.

\section{Experiment Setup}
We conduct the experiments on two widely used benchmark datasets: NYT10 and GIDS (Google-IIsc Distant Supervision dataset). The source code is publicly available\footnote{\href{https://github.com/nju-websoft/FAN}{https://github.com/nju-websoft/FAN}}.

\subsection{Datasets}
The statistics of two datasets are listed in Table \ref{tab:dataset}. We briefly describe them below:

\begin{compactitem}
    \item \textbf{NYT10} is developed by \cite{riedel} through aligning Freebase with the New York Times corpus. News from year 2005 to 2006 are used for creating the training set and from year 2007 for the test set. The entity mentions are annotated using Stanford NER \cite{StanfordNER} and linked to Freebase. The dataset has been broadly used for RE \cite{Hoffmann, Surdeanu, RESIDE, DISTRE}.
    
    \item \textbf{GIDS} is built by extending the Google RE corpus\footnote{\href{https://research.googleblog.com/2013/04/50000-lessons-on-how-to-read-relation.html}{https://research.googleblog.com/2013/04/50000-lessons-on-how-to-read-relation.html}} with additional instances for each entity pair \cite{GIDS}. It assures that the \textit{at-least-one} assumption of multi-instance learning holds, which makes the automatic evaluation more accurate and reliable.
\end{compactitem}

\subsection{Comparative Models}
To evaluate the proposed FAN, we compare it with the following seven representative models:
\begin{compactitem}
    \item \textbf{PCNN-ONE} \cite{PCNN}: A CNN-based neural RE model using piecewise max pooling for better sentence representations.
    
    \item \textbf{PCNN-ATT} \cite{ATT}: A model using selective attention to choose useful information across different sentences in a bag.
    
    \item \textbf{BGWA} \cite{GIDS}: An attention-based neural model which formulates various word attention and entity attention mechanisms to help a RE model focus on the right context.
    
    \item \textbf{RESIDE} \cite{RESIDE}: A GCN-based model which uses side information from a KB to improve performance.
    
    \item \textbf{DISTRE} \cite{DISTRE}: A GPT-based model capturing semantic and syntactic information along with commonsense knowledge.
    
    \item \textbf{DS-GAN} \cite{GANDriven}: A GAN-based model which treats distantly supervised RE as a semi-supervised learning process.
    
    \item \textbf{DS-VAE} \cite{VAE}: A variational auto-encoder (VAE) based model which biases the latent space of sentences and is trained jointly with a relation classifier.
\end{compactitem}

\begin{table}
    \centering
    {\small
    \begin{tabular}{ccrr}
    \toprule
    Datasets & Splits & Sentences & Entity pairs \\
    \midrule
    \multirow{3}{*}{\shortstack{NYT10 \\ (Relations: 53)}} & Train & 455,771 & 233,064 \\ 
      & Dev & 114,317 & 58,635 \\
      & Test & 172,448 & 96,678 \\
    \midrule
    \multirow{3}{*}{\shortstack{GIDS \\ (Relations: 5)}} & Train & 11,297 & 6,498 \\
     & Dev & 1,864 & 1,082 \\
     & Test & 5,663 & 3,247 \\
    \bottomrule
    \end{tabular}}
    \caption{Statistics of NYT10 and GIDS.}
    \label{tab:dataset}
\end{table}

\subsection{Criteria}

\begin{figure*}
    \centering
    \subfigure[NYT10]{
        \includegraphics[width=\columnwidth]{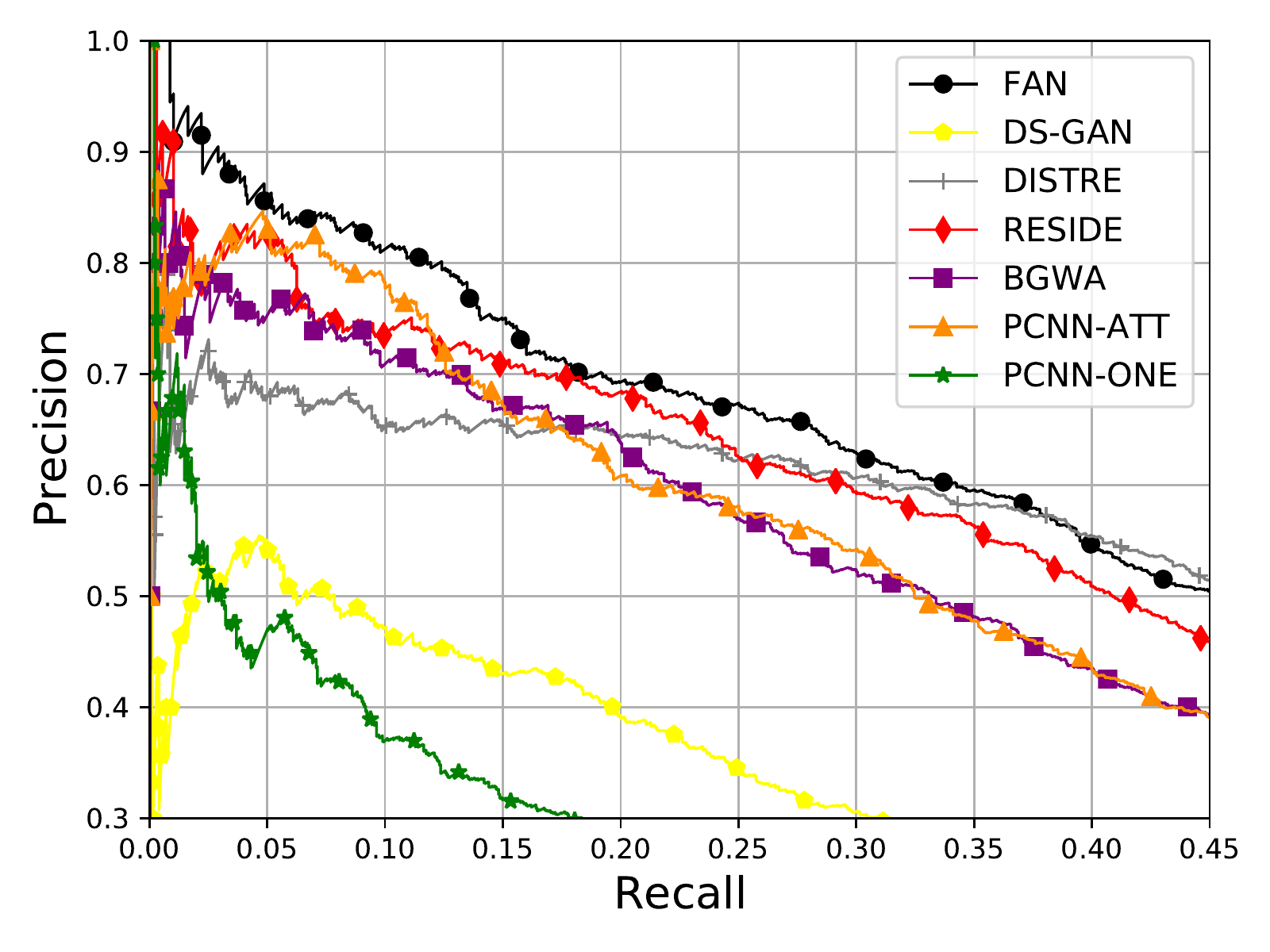}
        \label{fig:PR-NYT10}
    }
    \subfigure[GIDS]{
        \includegraphics[width=\columnwidth]{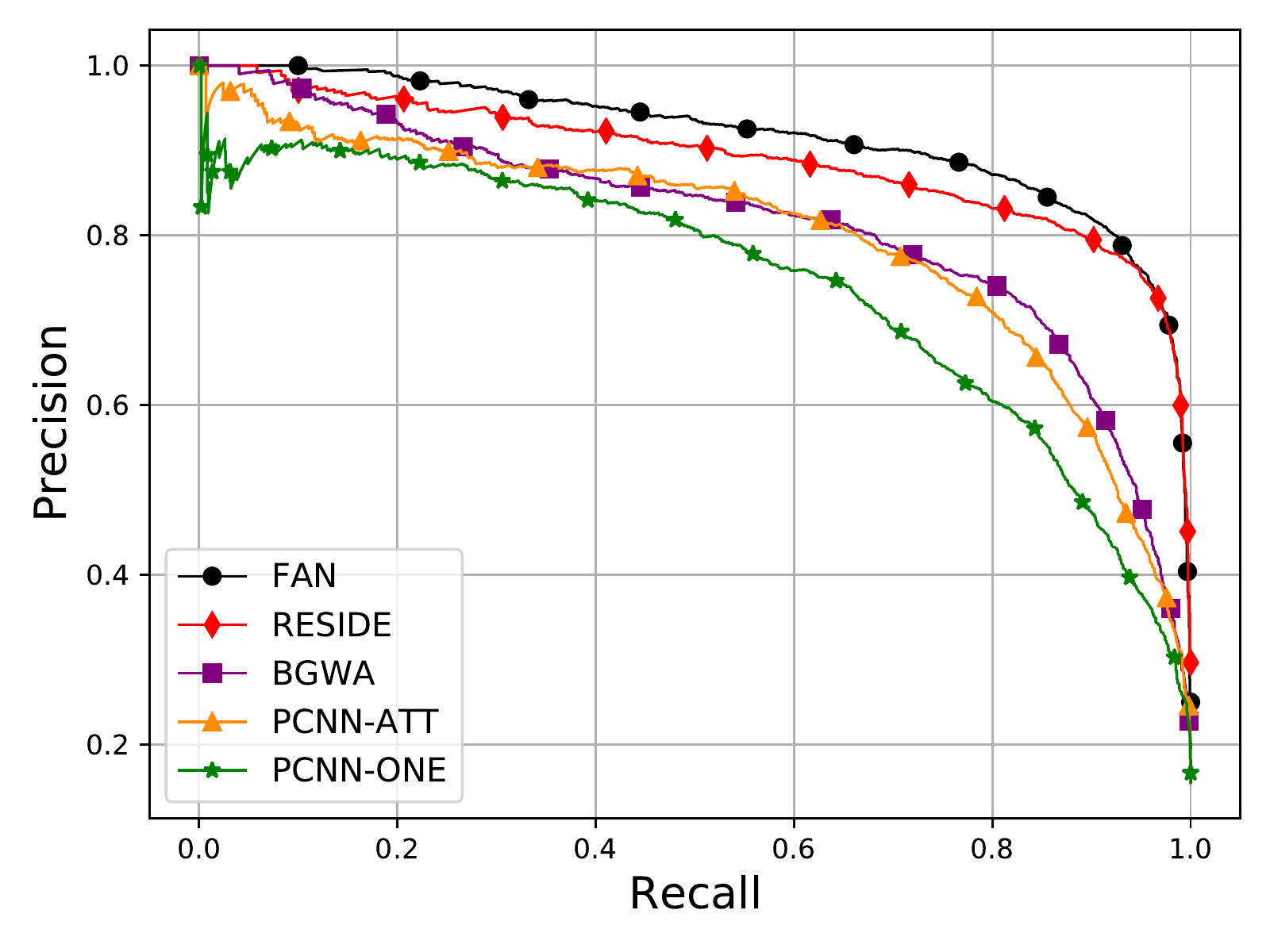}
        \label{fig:PR-GIDS}
    }
    \caption{PR-curve of different models. FAN achieves the state-of-the-art on both datasets.}
    \label{fig:PR}
\end{figure*}

Following the conventions \cite{PCNN,ATT,RESIDE,DISTRE}, we use the held-out evaluation. For each model, we compute the precision-recall (PR) curves and report the area under curve (AUC) scores for straightforward comparison. We also report P@N, which measures the percentage of correct classifications in the top-N most confident predictions. Additionally, micro-F1 is measured at different points along the PR-curve and the best is reported.

For NYT10, we compare with all models listed above. We reuse the results reported in the original papers for BGWA, RESIDE, DISTRE and DS-VAE, and implement PCNN-ONE, PCNN-ATT, DS-GAN by ourselves. Because the source code of DS-VAE is not released yet, we cannot obtain its PR-curve. For GIDS, we only compare with PCNN-ONE, PCNN-ATT, BGWA and RESIDE, as other models are not applicable to this dataset.

\section{Results}

\subsection{Overall Results}

The PR-curves on NYT10 and GIDS are shown in Figure \ref{fig:PR}. On both datasets, FAN achieves the best results. On NYT10, we get visibly higher recall, especially when precision is higher than 75.0, which indicates that our model can find more informative samples along with correct labels. It makes sense because FAN knows more information by digging informative samples from the N/A set and weakens improper biases in the training procedure. 
On NYT10, our AUC is $\textbf{45.5}$, improving $3.3$ on the basis of the second best model. 
On GIDS, our AUC is $\textbf{90.3}$, improving $1.2$ compared with the second best. 
Please refer to Table~\ref{tab:AUC} for details.

\begin{table}
    \centering
    {\small\begin{tabular}{lcc}
    \toprule
    & NYT10 & GIDS \\
    \midrule
    PCNN-ONE & 15.7 & 74.4 \\
    PCNN-ATT & 37.6 & 79.9 \\
    BGWA & 34.0 & 81.5 \\
    RESIDE & 41.5 & 89.1 \\
    DISTRE & 42.2 & - \\
    DS-GAN & 20.3 & - \\
    DS-VAE & 42.9 & - \\
    \midrule
    FAN & $\textbf{45.5}$ & $\textbf{90.3}$ \\
    \bottomrule
    \end{tabular}}
    \caption{AUC scores on NYT10 and GIDS.}
    \label{tab:AUC}
\end{table}

Table \ref{tab:P@N} shows P@N in top ranked samples and micro-F1 values on NYT10. Our model improves 3.8 on average for P@N, indicating that our model does not reduce the precision while improving the recall. Table \ref{tab:P@N4GIDS} shows the results on GIDS. From the table, we observe that FAN achieves the highest score in micro-F1. The scores of P@N are also comparable to RESIDE and BGWA.

\subsection{Mining Results}
In the mining step, we filter FN samples from $\mathcal{N}$ with logits larger than threshold $\theta$. As a result, we discover 4,556 FN samples from NYT10, which refer to 3,733 entity pairs; and 238 FN samples from GIDS, which refer to 225 entity pairs.

\begin{table}[!t]
    \centering
    \resizebox{\linewidth}{!}{
    \begin{tabular}{lcccc}
    \toprule
     & P@100 & P@200 & P@300 & Micro-F1 \\
    \midrule
    PCNN-ONE & 49.5 & 44.7 & 44.8 & 24.3 \\
    PCNN-ATT & 82.8 & 79.9 & 75.0 & 42.0 \\
    BGWA & 82.0 & 75.0 & 72.0 & 42.1 \\
    RESIDE & 81.8 & 75.4 & 74.3 & 45.7 \\
    DISTRE & 68.0 & 67.0 & 65.3 & 48.6 \\
    DS-GAN & 60.6 & 57.3 & 51.8 & 30.7 \\
    DS-VAE & 83.0 & 75.5 & 73.0 & - \\
    \midrule
    FAN & \textbf{85.8} & \textbf{83.4} & \textbf{79.9} & \textbf{48.7} \\
    \bottomrule
    \end{tabular}}
    \caption{P@N and micro-F1 on NYT10.}
    \label{tab:P@N}
\end{table}

\begin{table}[!t]
    \centering
    \resizebox{\linewidth}{!}{
    \begin{tabular}{lcccc}
    \toprule
     & P@100 & P@200 & P@300 & Micro-F1 \\
    \midrule
    PCNN-ONE & 88.1 & 90.5 & 90.0 & 70.0 \\
    PCNN-ATT & 97.0 & 93.5 & 91.4 & 75.6 \\
    BGWA & 99.0 & \textbf{98.0} & 96.0 & 77.3 \\
    RESIDE & \textbf{100.0} & 97.5 & \textbf{97.0} & 84.6 \\
    \midrule
    FAN & 98.3 & 97.6 & 96.8 & \textbf{85.9} \\
    \bottomrule
    \end{tabular}}
    \caption{P@N and micro-F1 on GIDS.}
    \label{tab:P@N4GIDS}
\end{table}

To evaluate the quality of $\mathcal{M}$, we choose five relations assigned by FAN with most samples, each selecting 100 sentences with highest confidence scores. Three well-trained NLP annotators are asked to annotate sentences in a binary way, to see whether the assigned pseudo labels are correct. The results are shown in Table \ref{tab:mining}. 
The average precision is 87.0, improving around 17.0 compared with the original NYT10 dataset \cite{riedel}. It verifies both the quality of the mined data and the effectiveness of the aligning step. The ``nationality'' relation gets relatively lower precision because for some sentences about sports, there are usually more than one person and one country mentioned, the model gets confused in this scenario.

\begin{table}
    \centering
    {\small
    \begin{tabular}{lc}
    \toprule
    Relations & P@100 \\
    \midrule
    /location/location/contains & 92.0 \\
    /people/person/place\_lived	& 87.0 \\
    /people/person/nationality & 74.0 \\
    /business/person/company & 94.0 \\
    /location/administrative\_division/country & 88.0 \\
    \midrule
    Average & 87.0 \\
    \bottomrule
    \end{tabular}}
    \caption{P@100 for top 5 relations in $\mathcal{M}$.}
    \label{tab:mining}
\end{table}

\subsection{Adversarial Domain Adaptation}
\begin{figure}
    \centering
    \subfigure[Before]{
        \includegraphics[width=0.465\columnwidth]{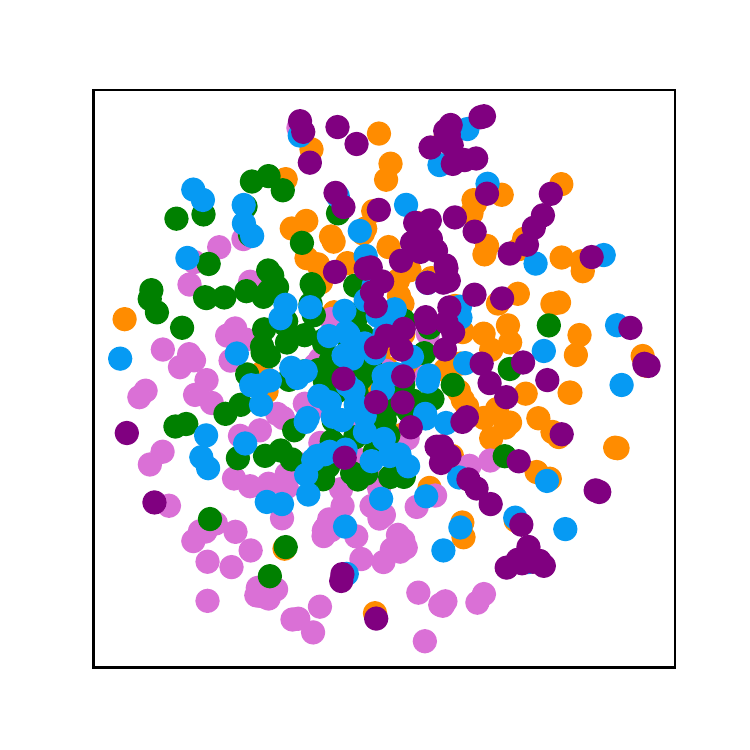}
        \label{fig:before}
    }
    \subfigure[After]{
        \includegraphics[width=0.465\columnwidth]{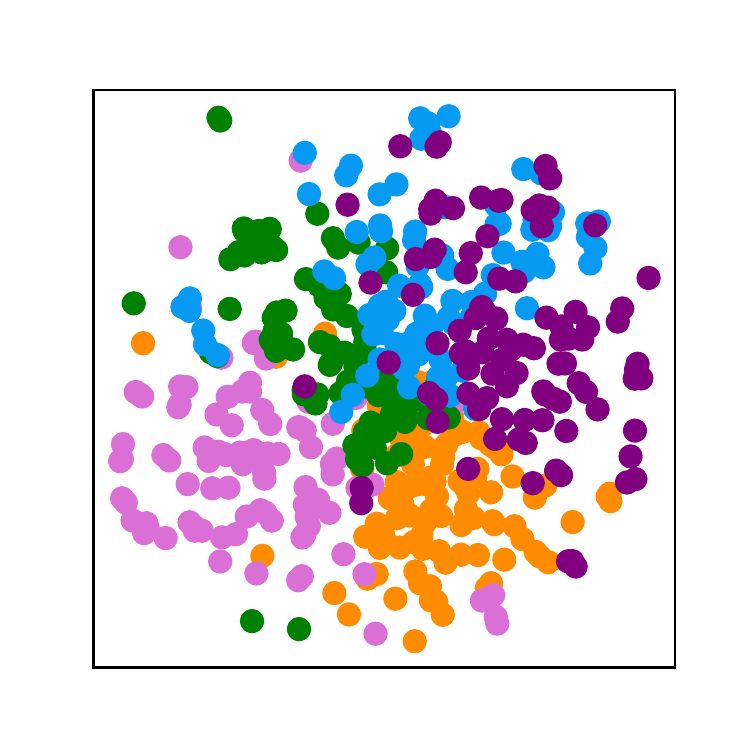}
        \label{fig:after}
    }
    \caption{T-SNE visualization of representations before and after aligning for $\mathcal{M}$.}
    \label{fig:DA}
\end{figure}

The label distributions may shift between $\mathcal{M}$ and $\mathcal{P}$. The generator aligns the two distributions into a unified space.
We use bag representations obtained through the bag encoder as the input of T-SNE to perform dimension reduction and obtain two-dimensional representations. As seen from Figure~\ref{fig:DA}, the feature distributions before aligning are overlapped and the classification boundary is not clear. After aligning, the samples are better clustered.

\subsection{Ablation Study}
We conduct an ablation study to verify the effectiveness of submodules in FAN. Table \ref{tab:ablation} shows the comparison results. (1) BERT has a great influence on the results because it introduces valuable linguistic knowledge and commonsense knowledge to RE. If replacing it by GloVe, both AUC and micro-F1 drop significantly. (2) By removing GRL and unlabeled data, using only the refined training set $\mathcal{D}^\prime$, both AUC and micro-F1 drops. This indicates that the information contained in unlabeled data is helpful for distantly supervised RE, and FAN can leverage it reasonably. (3) Contrastive loss can help the learning procedure. Different classes can be better clustered, thus reducing errors on the classification boundary.

\begin{table}
    \centering
    {\small
    \begin{tabular}{lrr}
    \toprule
     & AUC & Micro-F1 \\
    \midrule
    FAN & 45.5 & 48.7 \\
    \quad w/o BERT & 39.8 & 44.4 \\
    \quad w/o GRL & 43.7 & 47.1 \\
    \quad w/o contrastive & 44.7 & 48.0 \\
    \bottomrule
    \end{tabular}}
    \caption{Results of ablation study.}
    \label{tab:ablation}
\end{table}

\subsection{False Negatives in the Test Set}
\begin{figure}[!t]
    \centering
    \subfigure[AUC]{
        \includegraphics[width=0.46\columnwidth]{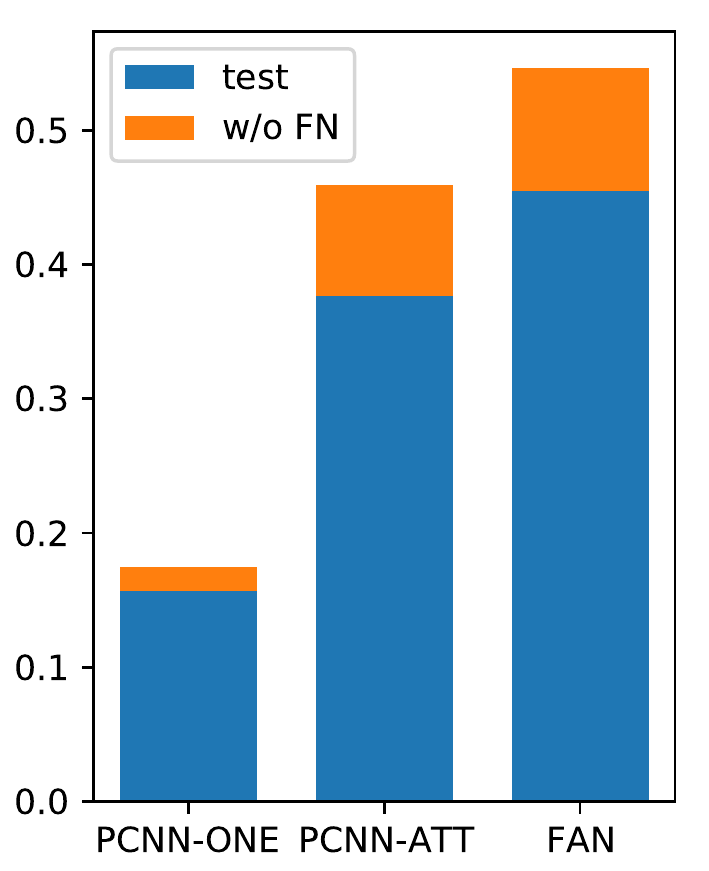}
        \label{fig:FN-test-AUC}
    }
    \subfigure[Micro-F1]{
        \includegraphics[width=0.46\columnwidth]{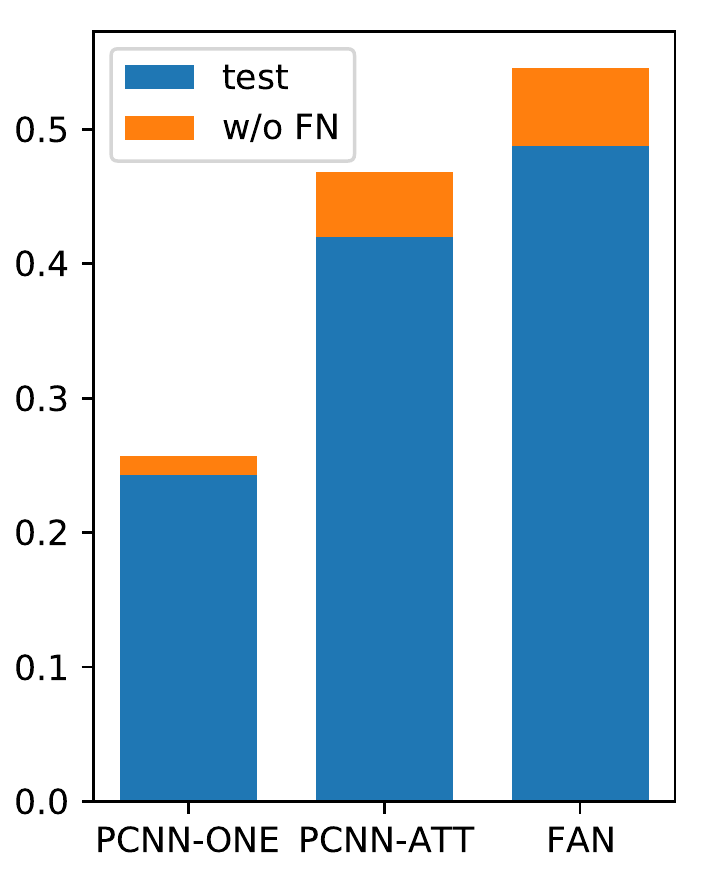}
        \label{fig:FN-test-Micro-F1}
    }
    \caption{Improvements of AUC and micro-F1 after removing FN in the test set.}
    \label{fig:FN-test}
\end{figure}

We also investigate the impact of FN on the test set during evaluation. As what we do in the training phase, we train a deep filter to mine FN from the test N/A set. As a result, we obtain 6,468 sentences, containing 4,951 entity pairs. By simply removing these data, we obtain huge improvements on AUC by around 20\% higher than the original to 54.6, and 12\% on micro-F1 to 54.5. See Figure \ref{fig:FN-test}. Similar phenomena occur on different baseline models.

This indicates that the FN samples in the N/A set greatly affect the evaluation procedure and bring in improper biases for model selection. For comparison, we randomly remove the same number of sentences from the test N/A set, as a result, AUC increases 0.4 and micro-F1 increases 0.9. This result is rational because randomly removing 6,468 samples is negligible in comparison with 166,004 N/A samples in the test set.

\subsection{Case Study}
In Table \ref{tab:case}, we show several cases of mined data which are excerpted from NYT10. 
\begin{compactenum}
\item The first sentence is correctly assigned with label ``/people/person/place\_lived'' by FAN, but it is missed by DS-GAN. Because one person is unusual to be mentioned on a Wikipedia page of a location, samples with relations between people and location are greatly omitted.

\item For less-frequent relations such as ``/sports/sports\_team/location'', FAN can still identify it to enlarge the training data and weaken the imbalance between relations.

\item In the third sentence, Queens contains Corona, but not reversely. FAN incorrectly assigns ``/location/location/contains'' from Corona to Queens. In fact, differentiating relation directions is a hard task and needs further study.
\end{compactenum}

\begin{table}
\centering
\resizebox{\columnwidth}{!}{
\begin{tabular}{llll}
	\hline
		\multicolumn{4}{|p{1.12\columnwidth}|}
		{Among those former clients are two wealthy sons of \textcolor{orange}{\textbf{[Robert C. McNair]}} , a \textcolor{orange}{\textbf{[Houston]}} oilman and the owner of the Houston Texans of the National Football League .} \\
	\hline 
		\multicolumn{4}{p{1.12\columnwidth}}
		{Predicted \& gold standard: \textcolor{cyan}{\textbf{\text{/people/person/place\_lived}}}} \\ & & & \\
	\hline
		\multicolumn{4}{|p{1.12\columnwidth}|}
		{... the mayor said , defending his advocacy of a new stadium on the Far West Side of \textcolor{orange}{\textbf{[Manhattan]}} for the 2012 Olympics and the \textcolor{orange}{\textbf{[New York Jets]}} .} \\
	\hline
		\multicolumn{4}{p{1.12\columnwidth}}
		{\resizebox{1.12\columnwidth}{!}{
		Predicted \& gold standard: \textcolor{cyan}{\textbf{\text{/sports/sports\_team/location}}}}} \\ & & & \\
	\hline
		\multicolumn{4}{|p{1.12\columnwidth}|}
		{The good news : Marta Caiza of \textcolor{orange}{\textbf{[Corona]}} , \textcolor{orange}{\textbf{[Queens]}} , won a new car .} \\
	\hline
	    \multicolumn{4}{p{1.12\columnwidth}}
	    {Predicted: \textcolor{magenta}{\textbf{\text{/location/location/contains}}}} \\
	    \multicolumn{4}{p{1.12\columnwidth}}
	    {Gold standard: \textcolor{magenta}{\textbf{\text{N/A}}}}
\end{tabular}}
\caption{Case study of mined data. \textcolor{cyan}{\textbf{correct prediction}} and \textcolor{magenta}{\textbf{wrong prediction}} are colored accordingly.}
\label{tab:case} 
\end{table}

\section{Conclusion and Future Work}
In this paper, we propose FAN, a two-stage method using adversarial DA to handle the FN problem in distantly supervised RE. We mine FN samples using the memory mechanism of deep neural networks. We use GRL to align unlabeled data with training data and generate pseudo labels to correct improper biases in both training and testing procedures. Our experiments show the superiority of FAN against many comparative models. In future work, we plan to use the teacher-student model to deal with FP and FN simultaneously.

\section*{Acknowledgements}
This work was supported by the National Natural Science Foundation of China (No. 61872172), and the Water Resource Science \& Technology Project of Jiangsu Province (No. 2019046).

\bibliographystyle{acl_natbib}
\bibliography{custom}

\clearpage
\balance

\appendix




\section{Experiment Setup}

In this section, we provide more details of our experiments. We implement FAN with PyTorch 1.6 and train it on a server with an Intel Xeon Gold 5117 CPU, 120GB memory, two NVIDIA Tesla V100 GPU cards and Ubuntu 18.04 LTS. 

The parameters of FAN are initialized with Xavier \cite{Glorot2010UnderstandingTD} using a fixed initialization seed. We train FAN by SGD optimizer with mini-batch of size 160. In the convolutional module, we set kernel size in $\{2, 3, 4, 5\}$ and filter size to 230. For details, please refer to Table \ref{tab:hyperparameters}.

\begin{table}[!h]
    \centering
    {\small\begin{tabular}{l|c}
    \toprule
    Hyperparameters & Values \\
    \midrule
    Batch size & 160 \\
    Learning rate & 0.1 \\
    Optimizer & SGD \\
    Filter size & 230 \\
    Kernel size & $\{2,3,4,5\}$ \\
    Weight decay & 0.00001 \\
    $\alpha$ & 0.01 \\
    $\beta$ & 0.01 \\
    $\gamma$ & 0.0001 \\
    $\theta$ & 0.5 \\
    $\tau$ & 1.0 \\
    Epoch & 50 \\
    Dim. of word embeddings & 768 \\
    Dim. of position embeddings & 50 \\
    Dropout rate & 0.5 \\
    \bottomrule
    \end{tabular}}
    \caption{Hyperparameters of FAN.}
    \label{tab:hyperparameters}
\end{table}

\section{Dataset Availability}

The NYT10 dataset \cite{riedel} is available at \href{http://iesl.cs.umass.edu/riedel/ecml/}{http://iesl.cs.umass.edu/riedel/ecml/}. We use the version adapted by OpenNRE \cite{OpenNRE}, which has removed the overlapped samples between the training set and the test set. The data is available at \href{https://github.com/thunlp/OpenNRE}{https://github.com/thunlp/OpenNRE}. 
The GIDS dataset is available at \href{https://github.com/SharmisthaJat/RE-DS-Word-Attention-Models}{https://github.com/SharmisthaJat/RE-DS-Word-Attention-Models}.

\section{False Negatives in N/A}

In this section, we give several examples of FN samples mined from the training N/A set. Those samples are assigned with pseudo-labels and confidence scores by FAN. They are diverse and contain many relation types, which verifies that the noises in N/A is not negligible and deserves further study.

\begin{table}[!b]
\centering
\resizebox{\columnwidth}{!}{
\begin{tabular}{llll}
	\hline
		\multicolumn{4}{|p{1.12\columnwidth}|}
		{On the other hand , a recent lake study by researchers at \textcolor{orange}{\textbf{[Middlebury College]}} , in \textcolor{orange}{\textbf{[Vermont]}} , found nothing similar .} \\
	\hline 
		\multicolumn{4}{p{1.12\columnwidth}}
		{Pseudo-label: \textcolor{cyan}{\textbf{\text{/location/location/contains}}} \, $0.999$}\\ & & & \\
	
	\hline
		\multicolumn{4}{|p{1.12\columnwidth}|}
		{I need all the help I can get , said \textcolor{orange}{\textbf{[Chet Culver]}} , the Democratic secretary of state in \textcolor{orange}{\textbf{[Iowa]}} and a likely contender for governor in 2006 .} \\
	\hline
		\multicolumn{4}{p{1.12\columnwidth}}
		{Pseudo-label: \textcolor{cyan}{\textbf{\text{/people/person/place\_lived}}} \, $0.997$}\\ & & & \\

	\hline
		\multicolumn{4}{|p{1.12\columnwidth}|}
		{Meb Keflezighi , the American who won the silver medal in the men 's marathon in last year 's Athens Olympics , and \textcolor{orange}{\textbf{[Liu Xiang]}} of \textcolor{orange}{\textbf{[China]}} , who won the men 's 110-meter hurdles in Athens , were entered yesterday in track and field competitions in New York .} \\
	\hline
		\multicolumn{4}{p{1.12\columnwidth}}
		{Pseudo-label: \textcolor{cyan}{\textbf{\text{/people/person/nationality}}} \, $0.999$}\\ & & & \\
		
	\hline
		\multicolumn{4}{|p{1.12\columnwidth}|}
		{If they think we 're the weakest link , they 're truly underestimating this little airline , said \textcolor{orange}{\textbf{[John Denison]}} , the new chief executive at \textcolor{orange}{\textbf{[ATA Airlines]}} , which sought bankruptcy protection on Oct. 26 .} \\
	\hline
		\multicolumn{4}{p{1.12\columnwidth}}
		{Pseudo-label: \textcolor{cyan}{\textbf{\text{/business/person/company}}} \, $0.999$}\\ & & & \\
		
	\hline
		\multicolumn{4}{|p{1.12\columnwidth}|}
		{The campaign against this poor little church outside \textcolor{orange}{\textbf{[Hangzhou]}} , the capital of Zhejiang Province in eastern \textcolor{orange}{\textbf{[China]}} , is part of a national wave of repression against independent , or underground , churches that are not registered with the government and do not recognize the authority of state-appointed spiritual leaders .} \\
	\hline
		\multicolumn{4}{p{1.12\columnwidth}}
		{Pseudo-label: \textcolor{cyan}{\textbf{\text{/location/administrative\_division/country}}} \, $0.999$}\\ & & & \\

    \hline
		\multicolumn{4}{|p{1.12\columnwidth}|}
		{For investors seeking the high returns that are no longer possible in the mature European and North American real estate markets , India and China are hot , said Prakash Gurbaxani , the chief executive of TSI Ventures in Bangalore , a joint venture of Tishman Speyer Properties of New York and India 's largest privately owned bank , \textcolor{orange}{\textbf{[ICICI Bank]}} , based in \textcolor{orange}{\textbf{[Mumbai]}} .} \\
	\hline
		\multicolumn{4}{p{1.12\columnwidth}}
		{Pseudo-label: \textcolor{cyan}{\textbf{\text{/business/company/place\_founded}}} \, $0.971$}\\ & & & \\
		
	\hline
		\multicolumn{4}{|p{1.12\columnwidth}|}
		{\textcolor{orange}{\textbf{[Peter Benenson]}} was born in \textcolor{orange}{\textbf{[London]}} on July 31 , 1921 , the son of a British army colonel .} \\
	\hline
		\multicolumn{4}{p{1.12\columnwidth}}
		{Pseudo-label: \textcolor{cyan}{\textbf{\text{/people/person/place\_of\_birth}}} \, $0.823$}\\ & & & \\
		
\end{tabular}}
\caption{Examples of FN.}
\label{tab:morecase} 
\end{table}

\end{document}